\documentclass{article}
\usepackage{kr}
\usepackage{times}
\usepackage[english]{babel} 
\usepackage{amsmath,amsfonts,amsthm,amssymb,mathrsfs} 
\usepackage{enumerate}
\usepackage{enumitem}
\usepackage{tikz}
\renewcommand{\citeyear}{\shortcite}
\usetikzlibrary{calc}
\usepackage{times}
\usepackage{soul}
\usepackage{url}
\usepackage[utf8]{inputenc}
\usepackage[small]{caption}
\usepackage{graphicx}
\usepackage{amsmath}
\usepackage{amsthm}
\usepackage{booktabs}
\usepackage{algorithm}
\usepackage{algorithmic}
\def \D{\Delta}
\def \R{\mathbb{R}}
\def \X{\mathscr{X}}
\def \O{\mathscr{O}}

\def \L{\mathcal{L}}
\def \F{\mathcal{F}}
\def \M{\mathcal{M}}
\def \K{\mathcal{K}}
\def \A{\mathcal{A}}
\def \V{\mathcal{V}}
\def \I{\mathcal{I}}
\def \C{\mathscr{C}}
\def \P{\mathscr{P}}

\def \S{\mathscr{S}}
\def \PR{\mathscr{PR}}
\def \RS{\mathit{RS}}
\def \val{\pi}

\def \OO{D}

\def \<{\langle}
\def \>{\rangle}

\newcommand{\Consis}{\mathit{Consis}}
\renewcommand{\iff}{\Leftrightarrow}

\renewcommand{\implies}{\Rightarrow}
\newcommand{\rimp}{\Rightarrow}
\newcommand{\commentout}[1]{}
\newcommand{\journal}[1]{}
\newcommand{\Lbc}{\mathcal{L}^{\mathit{bc}}}

\newcommand{\rep}{\mathit{rep}}

\def \E{\mathbb{E}}

\def \W{S}
\def \w{s}
\newcommand{\LQXAn}{{\cal L}^{\forall,X,\ell,A}_n}
\newcommand{\LQXAi}{{\cal L}^{\forall,X,\ell,A}_i}
\newcommand{\LQXAnD}{{\cal L}^{\forall,X,\ell,A,*}_n}
\newcommand{\mi}{\ell}
\newcommand{\sat}{\models}
\newcommand{\intension}[2]{[\![ #1 ]\!]_{#2}}

\def \prop{\textsc{prop}}
\def \supp{\textsc{supp}}
\def \cons{\textsc{cons}}

\renewcommand{\phi}{\varphi}

\newcommand{\shortv}[1]{}

\newtheorem{example}{Example}

\newtheorem{definition}{Definition}
\newtheorem{assumption}{Assumption}

\newtheorem{remark}{Remark}

\begin{document}

\title{Dynamic Awareness}
\author{Joseph Y. Halpern and Evan Piermont}

\author{
Joseph Y. Halpern$^1$
\and
 Evan Piermont$^2$
\affiliations
$^1$Cornell University Computer Science Department\\
$^2$Royal Holloway, University of London, Department of Economics
\emails
halpern@cs.cornell.edu,
evan.piermont@rhul.ac.uk
}

\maketitle

\begin{abstract}
We investigate how to model the beliefs of an agent who becomes
  more aware. 
We use the framework of Halpern and R\^ego \citeyear{HR09}
by adding probability, and define a notion of a \emph{model
transition} that describes constraints on how, if an
agent becomes aware of a new formula $\phi$ in state $s$ of a model $M$,
she transitions to state $s^*$ in a model $M^*$.   We then discuss how
such a model can be applied to \emph{information disclosure}. 
 \end{abstract}

\section{Introduction}

Agents must sometimes take actions in situations that they do not
fully comprehend. 
Work in computer science, economics, and philosophy has sought to
capture this formally by considering agents who are \emph{unaware} of
some aspects of the world.  (See, e.g., \cite{FH,HR09,MR94,MR99,HMS03,BC09,Sil08}.) 
Most work on awareness thus far has focused on the static case, where awareness
does not change.  The focus of this paper is the dynamic case: how to 
how to model the beliefs of an agent who becomes \emph{more} aware. 

When there is no unawareness, the standard approach to dealing with 
beliefs is well understood; we update by conditioning.
However, it is far from clear how beliefs should change when an agent
becomes aware of new features.
For example, a mathematician who becomes aware of the Riemann
hypothesis, but learns nothing beyond that, may change his methods of
proof and his beliefs about what is true, despite not having learned that any
particular event has obtained.
%
When an agent becomes more aware, his entire (subjective) view 
may change.
In this paper, we propose a model for the dynamics of increased
awareness
for agents who are introspective, and so can
reason about their own and other's awareness (and lack of it).

We work with a probabilistic extension of the framework of Halpern and R\^ego \citeyear{HR09}
 where agents understand that they themselves may be unaware of some
 propositions, but cannot directly reason about or articulate such
 propositions. Instead the agents consider states (possible worlds)
  that contain \emph{shadow propositions}---proxies that represent an
 agent's vague conception of those propositions of which he is
 unaware. 
We define a notion of a \emph{model
transition} that describes constraints on how, if an
agent becomes aware of a new formula $\phi$ in state $s$ of a model $M$,
she transitions to state $s^*$ in a model $M^*$.
When an agent becomes aware of a formula that she was previously unaware of,
she associates the novel propositions with shadow propositions she
previously considered. The states of the updated model resemble the
original states except that some shadow propositions may be replaced by the
novel propositions contained in the newly discovered formula.  

The transition rule allows uncertainty to both increase (because the agent
can be uncertain about how to interpret the new propositions) and
decrease (because by making a discovery, the agent learns about his
own unawareness). 
This is in contrast to \emph{reverse 
 Bayesianism} which, roughly speaking, requires that
if the agent is aware of both $\psi$ and $\psi'$,  and 
becomes aware of
$\phi$, then his assessment regarding the relative likelihood of $\psi$
and $\psi'$ should not change \cite{karni2013reverse}. 
Karni and Vier\o's requirement does not seem appropriate for many situations of
interest, especially for introspective agents, who may believe that the mere existence of $\phi$
is itself informative about the world---so becoming aware of $\phi$
changes beliefs about other propositions. 

For example, suppose that agent $i$ believes it is very likely that
he is fully aware and observes $j$ taking questionable actions.
He might then reasonably believe that agent $j$ is not rational.
But then suppose that $i$ becomes aware of some fact  $\phi$ of which
he was not previously aware. 
Since he now acknowledges that he did not fully understand the
situation, $i$ considers it more likely that $j$ can rationalize 
his behavior.
Hence, simply by becoming aware of $\phi$, $i$ changes his assessment
of $j$ being rational.

We then discuss how such a model can be applied to \emph{information disclosure}. Many
economic models predict voluntary disclosure due to strategic concerns
(because no news is interpreted as bad news). For example, consider
restaurant health code inspections, which rate restaurants on a scale
of A-F. Surely, an A restaurant would display the rating in the
window, so a rational and fully aware patron who sees a restaurant
without a rating can assume that the restaurant does not have an A
rating. But then, B 
restaurants would want their rating to be known, rather than having
patrons wonder where in the B to F range they fall; hence B
restaurants will also post their ratings.  Continuing inductively,
it follows that all restaurants will disclose their ratings (except
possible those with an F).   Economists have used such analyses to
conclude that many types of financial disclosures will be made
voluntarily, so regulation is not needed.

However, when agents might be unaware of the rating system, this
analysis clearly fails. An 
agent unaware of the rating system can draw no conclusion whatsoever
from the lack of a rating.  
Since disclosure might not only provide information but also expand
 awareness, belief and behavior might change in subtle ways.  
For example, we show that whether a restaurateur discloses a rating is
determined by his beliefs about how  patrons will interpret the rating
 if they were unaware of the system.  In particular, if the
 restaurateur believes that patrons view a
posted grade as arising from a binary pass-fail system, then the B
restaurant might withhold its rating, despite B objectively being an
above-average rating. 

More generally, there is a growing experimental literature suggesting
that subjects often fail to properly reason about hypothetical or
counterfactual events, and thus deviate from the predictions of
rational (and omniscient) agents; see, for example,
\cite{esponda2014hypothetical,john2016hiding,cox2017status,jin2018no}. Such
deviations need not result from wholesale irrationality, but
could arise from subjects' unawareness of the relevant
counterfactuals, for example, being unaware of the actions an opponent
could have taken but did not. Our model provides a framework
for considering unawareness in a strategic setting, and in particular,
about the effect of agents' beliefs about other agents' reactions to
becoming more aware.  

The rest of this paper is organized as follows: Section \ref{sec:lang}
introduces the syntax and semantics of our static propositional logic.  
We extend this model to a dynamic environment in Section
\ref{sec:dynamics}. The key definition is of a
\emph{transition rule}, which captures how a model can change when
an agent becomes aware 
of a new formula. Section \ref{sec:info} shows how thinking in terms of
transition rules can be helpful in making predictions in economic
environments of information disclosure, where one agent can
strategically decide to make another agent aware of a new
formula. Finally, Section \ref{sec:conc} concludes.  

\section{A (Static) Propositional Logic of Awareness and Probabilistic
  Beliefs}  
 \label{sec:lang}

In this section, we introduce a propositional logic of awareness and
probability that essentially combines ideas from the propositional
logic of awareness introduced by Halpern and R\^ego \citeyear{HR09} (henceforth HR13)
and the logic of knowledge and probability  introduced by Fagin and
Halpern \citeyear{FH3} (henceforth FH94).    The syntax
of the logic is as follows: given a set $\{1, \ldots, n\}$ of
agents, formulas are formed by starting with a countably infinite set $\Phi =
\{p_1, p_2, \ldots\}$ of primitive propositions and a countably
infinite set  $\X$ 
of variables, and then closing off under
conjunction ($\land$), negation ($\neg$), the modal operators
$A_i, X_i$, $i = 1, \ldots, n$, \emph{likelihood formulas} of
the form $a_1 \mi_{i}(\phi_1) + \cdots + a_k \mi_{i}(\phi_k) > b,$
where $a_1, \ldots, a_k, b$ are rational numbers, and 
quantification, so that if $\phi$ is a formula, then so is
$\forall x \phi$.

Some comments on the syntax:  Following Fagin and Halpern
\citeyear{FH}, $X_i$ denotes explicit knowledge (or belief); $X_i\phi$
is true if, in addition to $\phi$ being true in all states that $i$
considers possible, $i$ is aware of $\phi$.  We capture awareness
using the $A_i$ modality; $A_i \phi$ means that $i$ is
aware of $\phi$.  Awareness is 
syntactic, so $i$ may be aware of $p_1$ but not aware of $p_1 \land (p_2 \lor
\neg p_2)$.  
In this paper, we assume for simplicity that awareness
is \emph{generated by primitive propositions} (an assumption that goes
back to Fagin and Halpern \citeyear{FH}), so that an agent is aware of
a formula iff he is aware of all the primitive propositions in the
formula).  Following HR13,
we use quantification to capture knowledge of 
unawareness.  For example, the formula $X_i (\exists x \neg A_i(x))$
says that agent $i$ (explicitly) knows that there is a formula that he
is unaware of.  (As usual, $\exists x \phi$ is an abbreviation for
$\neg \forall x \neg \phi$.)
The $\mi_i$ in a likelihood formula can be interpreted as
probability, so a formula such as 
$a_1 \mi_{i}(\phi_1) + a_2 \mi_{i}(\phi_2) > b$ says that $a_1$ times
the probability of $\phi_1$ plus $a_2$ times the probability of
$\phi_2$ is at least $b$.    Call this language $\LQXAn(\Phi)$.

As in first-order logic, we can define
inductively what it means for a variable $x$ to be {\it
free} in a formula $\varphi$. Intuitively, an occurrence of a variable
is free in a formula if it is not bound by a quantifier. A formula
that contains no free variables is called a {\it sentence}.
If $\psi$ is a formula, let $\varphi[x/\psi]$ denote 
the formula that results by replacing
all free occurrences of the variable $x$ in $\phi$ by $\psi$.
(If there is no free occurrence of $x$ in $\varphi$, then
$\varphi[x/\psi]=\varphi$.) 
Unlike standard 
quantified modal logic, where the quantifiers range
over propositions (intuitively, sets of states), following 
HR13,
 here the quantifiers 
range over quantifier-free sentences.  Thus, $\forall x \phi$ is true
iff $\phi[x/\psi]$ is true for all quantifier-free sentences $\psi$.  
Roughly speaking, we want quantification to range over formulas, since
$A_i$ is syntactic.  However, it cannot range over all formulas,
since, for example, the formula $\forall x (x)$ would then be true iff all
formulas (including itself) were true, and 
we would not be able to get a recursive definition of truth.  We avoid
these difficulties by taking the domain of quantification to be the
quantifier-free sentences.  (See HR13 for further discussion
of these syntactic constraints.)

We give semantics to these formulas in probabilistic awareness
structures.  Given $\Phi$, let $\Phi^+$ consist of $\Phi$ together
with an infinite set $\Phi' = \{q_1, q_2, \ldots\}$ (disjoint from $\Phi$) of
special primitive propositions that we call \emph{shadow
  propositions}; we explain their role below.
A \emph{probabilistic awareness structure for $n$ agents (over
  $\Phi^+$)} is a tuple $M =(\W$, $\val$, $\K_1$, $\ldots$, $\K_n$,
  $\A_1$, $\dots$, $\A_n, \PR_1, \ldots, \PR_n,\L)$ satisfying the
  following properties:
\begin{itemize}
\item $\W$ is a set
  of states (or possible worlds), which we take to be finite for simplicity.
\item $\val: \W \times \Phi^+ \rightarrow \{{\bf true},{\bf false}\}$
  determines which primitive propositions in
$\Phi^+$ are true at each state in $\W$.
\item ${\cal K}_i$ is a binary relation on $\W$ for
each agent $i = 1, \ldots, n$, which for this paper we take to be
\emph{Euclidean} (if $(\w_1,\w_2) \in \K_i$ and
$(\w_1,\w_3) \in \K_i$, then $(\w_2,\w_3) \in \K_i$), 
transitive, and \emph{serial} (for all states $\w$, there exists a
state $\w'$ such that $(\w,\w') \in \K_i$). 
\item $\PR_i$ associates with each state $\w$ a probability
on the states in $\K_i(\w) = \{\w': (\w,\w') \in
\K_i\}$, where all subsets of $\K_i(\w)$ are taken to be measurable
and $\PR_i(\w') = \PR_i(\w)$ if $\w' \in
\K_i(\w)$. 
\item ${\cal A}_i$ is a function
associating a set of propositions with each state in $\W$, for $i=
1,...,n$ such that if $\w' \in \K_i(\w)$, then $\A_i(\w) =
\A_i(\w')$.%
\item $\L$ associates with each state $\w$ a subset of $\Phi^+$; we
    require that $(\cup_{i=1}^n \A_i(\w)) \subseteq \L(\w)$
and that if $\w' \in \K_i(\w)$,
then $\L(\w') \subseteq \A_i(\w) \cup \Phi'$.
\end{itemize}
Intuitively, if $(\w,\w') \in \K_i$, then agent $i$ considers
state $\w'$ possible at $\w$.  The assumption that
$\K_i$ is Euclidean, transitive, and serial means 
that the $K_i$ operator satisfies the axioms of KD45 traditionally associated
with the logic of belief (see \cite{FHMV} for more discussion).%
\footnote{As is often done, we blur the distinction between knowledge
  and belief in this discussion.}
$\A_i(s)$ is the set of primitive propositions 
that agent $i$ is aware of at state $\w$.
It is more standard to take $\A_i(\w)$ to be the set of \emph{all} formulas
that $i$ is aware of at world $\w$.  We are assuming here that
awareness is generated by primitive propositions \cite{FH}; agent $i$
is aware of a formula $\phi$ if $i$ is aware of all the primitive
propositions in $\phi$.  Thus, it suffices to take $\A_i(\w)$ to
consist only of primitive propositions.
The fact that $\A_i(\w') =
\A(\w)$ if $\w' \in \K_i(\w)$ means that an agent knows what he is
aware of; similarly, the assumption that 
$\PR(\w') = \PR(\w)$ if $\w' \in \K_i(\w)$ means that an agent knows
his beliefs. 


Intuitively, $\L(\w)$ is the
language associated with state $\w$.  We certainly want the language
to  include all the primitive propositions that some agent is aware of
at state $\w$, but possibly others as well.
For example, if agent $i$ knows at state $\w$ that
there is a formula that no agent is aware of (which can be expressed as $X_i
(\exists x (\neg A_1 (x) \land \ldots \land \neg A_n(x))$), then 
at each state $\w'$ that he considers possible,  the language must include a
primitive proposition not in $\cup_{j=1}^n \A_j(\w)$.  Moreover, as
noted by
HR13,
we must allow different
languages at different worlds to deal with the possibility that agent 
$i$ might be unsure of whether he is aware of all formulas (which can
be expressed as $\neg X_i \neg (\exists x \neg A_i(x)) \land \neg X_i
\neg (\forall x (A_i(x)))$).  In this case, there must be states $\w'$
and $\w''$ in $\K_i(\w)$ such that $\L(\w') = \A_i(\w)$ and $\L(\w'')
\supset \A_i(\w)$ (where $\supset$ denotes strict superset).   If $\w'
\in \K_i(\w)$, then we think of the primitive propositions in $\L(\w')
- \A_i(\w)$ as ``shadow propositions''.  Agent $i$ understands that
they must exist, at some level, but since $i$ is not aware of them, he
does not really understand what they denote. 
Thus, we require that $\L(\w')$ consist
of the ``real'' primitive propositions that $i$
is aware of and possibly some shadow propositions.  HR13
did not distinguish real and shadow
propositions.  Making the distinction has no effect on their axioms
involving awareness,
but we find it conceptually useful, especially to 
discuss belief dynamics in the next section.  However,
it is worth noting that this requirement means that, in general,
$\K_i$ will not be reflexive.  If $\L(\w)$ contains real primitive
propositions that agent $i$ is unaware of, then we cannot have $\w \in
\K_i(\w)$.  

We now define what it means for a formula to be true at a state $\w$ in a
probabilistic awareness structure $M$ by combining the earlier
definitions of FH94 and HR13.
  We take  $\LQXAn(\w)$ to consist of the formulas
  all of whose primitive propositions are in $\L(\w)$.
  For a formula $\phi$ let $\prop(\phi)$ denote the set of primitive
propositions in the formula $\phi$. 

\begin{itemize}
\item $(M,\w)\sat p$ if $p\in\L(\w)$ and $\val(\w,p)= {\bf true}$;
\item $(M,\w)\sat \neg \phi$ if $\phi\in \LQXAn(\w))$ and
$(M,\w)\not\sat \phi$;
\item $(M,\w)\sat \phi \land \psi$ if $(M,\w) \sat \phi$ and $(M,\w)
  \sat \psi$;
 \item $(M,\w)\sat A_i\phi$  if  $\prop({\phi}) \in \A_i(\w)$;
\item $(M,\w)\sat X_i\phi$  if $(M,\w)\sat A_i\phi$ and $(M,\w')
  \sat \phi$ for all $\w' \in \K_i(\w)$;
\item $(M,\w)\sat \forall x\varphi$ if $(M,\w) \sat \phi[x/\psi]$ for
all quantifier-free sentences $\psi \in \LQXAn(\w)$;
\item $(M,\w)\sat a_1 \mi_i(\phi_1) + \cdots + a_k \mi_i(\phi_k) > b$ 
if $a_1 \PR_i(\w)(\intension{\phi_1}{M} \cap \K_i(\w)) + \cdots + a_k
\PR_i(\w)(\intension{\phi_k}{M} \cap \K_i(\w)) > b$, where $\intension{\phi}{M} = \{\w':
(M,\w') \sat \phi\}$,
and $(M,\w) \sat A_i (\phi_1 \land \ldots \land \phi_k)$.
\end{itemize}

\begin{example}
\label{ex:jaware}
Consider a model with two agents, $i$ and $j$, and three states, $\w_1$, $\w_2$,
and $\w_3$. $\L(\w_1)$ consists of two propositions, $p$ and $p'$, both
``real''. In state $\w_1$, agent $i$ is 
unaware of $p'$, while  $j$ is aware of both. $\L(\w_2) = \{p,q\}$, where
$q$ is a shadow proposition.  In state $\w_2$, agent $j$
is aware of both $p$ and $q$, while $i$ is aware only of $p$. Finally,
$\L(\w_3) = \{p\}$, and both agents are aware of $p$ in state $\w_3$.
$\K_i(\w_k) = \{\w_2, \w_3\}$ and $K_j(\w_k) = \w_k$ for $k = 1,2,3$.  

In state $\w_1$, $i$ believes its possible that $j$ is aware of something
he himself is not aware of. However, he can only describe this state
vaguely, envisaging not $p'$, but the shadow proposition $q$. He also
considers it possible that he is fully aware. If $p$ is true
at state $s_2$, then $i$ knows that if he is unaware of
something, then $p$ is true. 
\qed
\end{example}

\section{Dynamics}
\label{sec:dynamics}

We now turn our attention to how an agent's beliefs should be updated
when the agent becomes aware of a formula $\phi$ of which he was previously
unaware.%
\footnote{Of course, an agent may become aware of $\phi$ by learning
$\phi$ (i.e., by learning that $\phi$ is true). 
 We view this conceptually as the composition of two updates:
the update due to becoming aware of $\phi$, followed by the update due
to learning $\phi$ given that the agent is aware of $\phi$, which can
be handled by standard techniques, specifically,
conditioning.}
As the general case is notationally complex, to ease exposition, we
first consider the single-agent case.
With only a single agent, the main concern is how the agent
interprets the formula he becomes aware of, and how knowledge and
probabilistic reasoning depend on his interpretation.
In the general case, considered in the next subsection, we also need
 to deal with the beliefs of other agents, and with higher-order beliefs.  

 \subsection{Dynamics: The Single-Agent Case}

Let $\LQXAi(\Phi)$ denote the language with a single agent $i$.
Suppose that in state $\w$ in a model $M = (\W$, $\val$, $\K_i$,
$\A_i$, $\PR_i$, $\L)$,  the (single)  
agent $i$ becomes aware of a formula $\phi$ of which he was previously
unaware.  
Our goal is to construct an updated model
$M^* = (\W^*$, $\val^*$, $\K_i^*$,
  $\A_i^*$, $\PR_i^*$,$\L^*)$  
  and state $\w^* \in \W^*$ 
  that reflects this.   
Each state in 
$M^*$ corresponds to some state in $M$; the correspondence is captured
by a relation $T$.  We now explain how $T$ works.


If $t$ is a state such that $\L(t)$ has at least as many
shadow propositions that $i$ is unaware of as there are 
propositions in $\phi$ that $i$ is unaware of, then $t$ is
\emph{consistent with $i$ 
  becoming aware of $\phi$}.  Let $\cons(M,\phi,i) \subseteq \W$ denote the
states consistent with $i$ becoming aware of $\phi$ in model $M$.
%
If $\PR_i(\w)(\cons(M,\phi,i))  = 0$, then
becoming aware of $\phi$ was an 
event to which $i$ previously gave probability 0; $i$ didn't consider
it possible that there were two primitive propositions of which he
was unaware.  In this case, we place no constraints on how $i$ updates
his beliefs; this is analogous to conditioning on an event of measure
0.   (We make precise below what ``no constraints'' means%
.)  

If, on the other hand, $\PR_i(\w) (\cons(M,\phi,i)) \ne 0$,
then for each $t \in \K_i(\w) \cap \cons(M,\phi,i)$, $i$
must decide how to take $\phi$ 
into account at $t$.  For example, suppose that at a state $s$ where
$i$ is unaware of $p$ and $p'$, $i$ becomes aware of the formula $p
\land p'$.  Further suppose that at $t \in \K_i(s)$, there are exactly three shadow
propositions, $q$, $q'$, and 
$q''$, in $\L(t)$.  Then $i$ must decide 
which of $q$, $q'$, and $q''$ is $p$ and which is $p'$; $i$ could
in principle decide that none of them is an
appropriate candidate for $p$ (or $p'$).  
%
For example, after becoming aware of $p \land p'$, $i$ might consider possible
a state $t^*$ such that (1)
$\L^*(t^*) = \L(t) - \{q,q'\} 
\cup \{p,p'\}$; (2)  $\A_i^*(t^*) = \A_i(\w) \cup \{p,p'\}$
(3)
$\val^*(t^*,r) = \val(t,r)$ if $r \in \L(t) - \{q,q'\}$, 
$\val^*(t^*,p) = \val(t,q)$, and $\val^*(t^*,p') = \val(t,q')$.
Thus, in $t^*$, $i$ has replaced the shadow proposition $q \in \L(t)$
with $p$, and $q'$ with $p'$.    Agent $i$ might also consider
possible a state where $q'$ is replaced by $p$ and $q''$ is replaced
by $p'$.  
Roughly speaking, in moving from $M$ to $M^*$, we want to replace
each state $t \in \K_i(\w)$ by some states compatible with $t$
(each of these states will be related to $t$ by $T$) and
distribute the probability of $t$ among these states, and then
condition on $\cons(M,\phi,i)$.

The next definition makes precise the relationship between related
states.  

\begin{definition}
    $f: \Phi^+ \to \Phi^+$ is a \emph{$\phi$-replacement scheme} 
if $f$ is the identity on $\Phi^+ - \prop(\phi)$ (where
$\prop(\phi)$ is the set of propositions in $\phi$) and $f$ is injective on
$\prop(\phi)$.  A $\phi$-replacement scheme $f$ is \emph{compatible
    with $i$ becoming aware of $\phi$ at 
  $\w$} if $f$ is the identity on $\prop(\phi) \cap
\A_i(s)$ and $f(\prop(\phi) - \A_i(\w)) \subseteq (L(\w) \cap \Phi') -
\A_i(\w)$. 
A state $\w^*$ is an \emph{$f$-replacement for $\w$} 
if 
(1) $f(\L^*(\w^*)) = \L(\w)$ and
(2) 
$\val^*(\w^*,p) = \val(\w,f(p))$ for all $p \in \L(\w^*)$.
\end{definition}
Intuitively, a $\phi$-replacement scheme for $i$
describes how $i$ 
interprets the propositions in $\phi$ of which he was previously
unaware, associating each with a different
(shadow) proposition.  That is, $f$ maps each proposition in $\phi$
that $i$ becomes aware of
to some (unique) shadow proposition that $i$ was unaware of,
and leaves all other propositions alone.  

If $f(p) = q$, then $i$
considers it possible that the shadow proposition $q$ represents the real
proposition $p$ that appears in $\phi$.
So as discussed above, if after becoming aware of $p \land p'$, $i$ considers $t^*$ where $p$ was represented by $q \in \L(t)$
and $p'$ by $q'$, then $t^*$ is a $f$-replacement of $t$ where $f(p) = q$, $f(p') = q'$, and $f$
is the identity otherwise. 
If $\PR_i(s)(\cons(M,\phi,i)) \ne 0$, then
the relation $T$ mentioned above actually associates each state $t^* \in
\K_i^*(s^*)$ with a pair $(t,f)$, where $t \in \K_i(s)$ and $t^*$ is
an  $f$-replacement of $t$ compatible with $i$ becmoing aware of
$\phi$ at $s$.





A \emph{situation} is a pair
$(M,s)$ consisting of a model $M$ and a state $s$ in $M$.  Fix a set
$\Phi$ of primitive propositions and a set $\I$ of agents.%
\footnote{Although this section deals only with the single-agent case,
  so $\I = \{i\}$, we introduce the more general notation for
  so that we can use it in later sections.}
Let
$\M(\I,\Phi)$ and $\S(\I,\Phi)$ consist of all models and situations,
respectively,  over the language $\Phi$ with agents in $\I$.  Let
$\RS(\phi,i,\w)$ denote the set of $\phi$-replacement schemes compatible
with $i$ becoming aware of $\phi$ at $\w$, and
let $\RS(\phi) = \{id\} \cup (\cup_{i,\w}R(\phi,i,w))$ (where $id$ is the
identity function 
on $\Phi^+$).

The next definition is the one that we have been heading for.  It
describes what counts as an acceptable update from a situation $(M,s)$
to a new situation  $(M^*,s^*)$ that is the result of agent $i$
becoming aware of $\phi$.  We capture this using the notion of an
acceptable transition rule $\tau$.  Formally,
a \emph{transition rule} $\tau$ maps a situation, %
a formula, and an
agent 
(for now, necessarily the single agent $i$)
to a new situation, interpreted as the
result of agent $i$ becoming aware of the formula $\phi$
when the initial situation was $(M,s)$.  $(M^*,s^*)$ is an acceptable
update from a situation $(M,s)$ after agent $i$ becomes aware of
$\phi$ if $\tau((M,s),i,\phi) = (M^*,s^*)$ for some acceptable
transition rule $\tau$.
Although, conceptually, the ideas behind the
definition are quite straightforward, writing them down carefully
results in a complicated definition.  We give some intuition for the
details of the definition immediately after the definition, and then
provide an example that illustrates some of them.

\begin{definition}
\label{def:single}
$\tau: \S(\{i\},\Phi) \times \LQXAi(\Phi) \times \{i\} \to  
\S(\{i\},\Phi)$ is an \emph{acceptable transition rule for a single agent} if for all
$(M,s) \in \S(\{i\},\Phi)$,
$\phi \in \LQXAi(\Phi)$ if
$\prop(\phi)\in \L(s)$,
$M = (\W, \val, \K_i, \A_i, \PR_i,\L)$, 
  $\tau((M,\w),\phi, i) = (M^*,\w^*)$, and
  $M^* = (\W^*, \val^*, \K_1^*, \A_1^*, \PR_1^*,\L^*)$,
  then either $\prop(\phi) \subseteq \A_i(s)$ and $(M,s)=(M^*,\w^*)$, or
  $\prop(\phi) \not\subseteq \A_i(s)$ and 
  there exists a relation $T \subseteq \W \times \W^* \times \RS(\phi)$
    such that the following hold:

    \begin{description}
\item[{\rm T1.}]
  For all $t^* \in \W^*$, there exists a unique $t \in \W$ and $f$ such
that $(t,t^*, f) \in T$; moreover,
 $t^*$ is an $f$-replacement of $t$. 
\item[{\rm T2.}] $(s,s^*,id) \in T$ and $\A^*_i(\w^*) = \A_i(\w) \cup
\prop(\phi)$.
\item[{\rm T3.}]  
  If $\PR_i(\w)(\cons(M,\phi,i)) \ne 0$, then for all $(t,t^*,f) \in T$ the following
  conditions hold:
    \begin{description}
  \item[{\rm (a)}] $\K_i^*(t^*) = \{t^{\dag}: (t',t^{\dag},f') \in
T, \mbox{ for some } t' \in \K_i(\w) \cap \cons(M,\phi,i), f' \in \RS(\phi,i,\w)\}$. 
      \item[{\rm (b)}] If $t' \in \K_i(\w) \cap \cons(M,\phi,i)$, let
                $\rep(t') = \{t^{\dag} \in \K_i^*(t'): \exists f' 
                ((t',t^{\dag},f') \in T)\}$.  Then  
        $\PR_i^*(t^*)(\rep(t')) =
        \PR_i(\w)(t')/\PR_i(\w)(\cons(M,\phi,i))$. 
\end{description}
\end{description}

\end{definition}



Note that the conditions in the definition apply only if
$\prop(\phi) \subseteq \L(s)$. State $s$ cannot be the actual world if
its language does not include propositions that appear in a formula
that is part of the description of the world.  We do not care what
$\tau$ does at situations $(M,s)$ that do not describe the world.
For situations that describe the world, nothing changes if $i$ was already
aware of all the primitive propositions in $\phi$.  If $i$ was not
aware of all the primitive propositions in $\phi$, then we
use the relation $T$ to talk about corresponding states.  T1--T3
describe the key properties of the $T$ relation.
T1 says that, in
$M^*$, each state $t^*$ is an $f$-replacement of some state $t$ in $S$.
T2 ensures that the
distinguished state $\w^*$ comes from $\w$ and, agrees with $\w$ as far
as the language and truth of primitive propositions goes (since it is
an $id$-replacement of $s$); moreover, $i$'s awareness changes
appropriately (since 
$i$ becomes aware of $\phi$).


The most interesting requirement is T3, which 
captures how $i$'s beliefs change in states in $\K_i^*(\w^*)$. Notice
that, by T2 and the 
constraints on awareness sets, 
$\A^*_i(t^*) = \A_i(\w) \cup \prop(\phi)$ for all states that $i$
considers possible at $s^*$. 
Thus, $i$'s awareness must be updated by adding the new propositions
in $\phi$, and he knows this has occurred. There are no further
constraints if $\PR_i(s)(\cons(M,\phi,i)) = 0$; we have
nothing to say about how $i$'s beliefs change if $i$ ascribes
probability 0 at state $s$ to the possibility of becoming aware of the new
primitive propositions in $\phi$.

If $i$ ascribed positive probability to
$\cons(M,\phi,i)$, then each state  $t' \in K_i(\w) \cap
\cons(M,\phi,i)$ corresponds to some states $t^{\dag}$ in $\K_i(\w^*)$;
each such state $t^{\dag}$ is a possible replacement of $t$, for some
$f \in \RS(\phi,i,\w)$, so $f$ is compatible with $i$ becoming
aware of 
$\phi$ at $\w$.

The probability of this set of replacements of $t'$, which
we denote $\rep(t')$, according to $\PR_i^*(t^*)$ 
(which is the same as $\PR^*_i(s^*)$) is
exactly the probability of $t'$ according to $\PR_i(s)$, conditional on
$\cons(M,i,\phi)$. 
Thus, $i$'s beliefs about propositions he
becomes aware of depend on his interpretation of the shadow
propositions.
This updating rule is explored in a
probability theoretic setting by \cite{piermont2019unforeseen}, 
where the primitive is a prior and a posterior measure, $\PR_i$ and
$\PR^*_i$, and where $\PR^*_i$ can be defined over a possibly richer
algebra of events.  
\citeauthor{piermont2019unforeseen} provides conditions on
this pair of measures that suffice to ensure that they arise via an
updating rule as 
given by T3(c).  

\begin{figure*}[]
\centering
\begin{tikzpicture}[scale=.8, font = \scriptsize,every text node part/.style={align=center}]
\def\x{2};
\def\y{1.6};
\def\e{.1};
\def\z{2};

\draw[rounded corners, thick,]  (1.5,0) rectangle (1.5+\x,0+\y)
node[pos=.5] {$p$ \\ $A_i = \emptyset$};
\node at (1.5+\x*.5,0-2*\e) {$s$};

\draw[rounded corners, thick,]  (1.5,2.5) rectangle (1.5+\x,2.5+\y) 
node[pos=.5] {$q,\neg q'$ \\ $A_i = \emptyset$};
\node[scale=2, color=red] at (1.5+\x-3*\e,2.5+\y-3*\e) {$\circlearrowleft$};
\node at (1.5+\x*.5,2.5-2*\e) {$t$};

\draw[rounded corners, thick,]  (0+3*\z,2.5) rectangle (0+3*\z+\x,2.5+\y) 
node[pos=.5] {$p,\neg q'$ \\ $A_i = \{p\}$};
\node[scale=2, color=red] at (0+\x-3*\e+3*\z,2.5+\y-3*\e) {$\circlearrowleft$};
\node at (0+\x*.5+3*\z,2.5-2*\e) {$t^*$};

\draw[rounded corners, thick,]  (3+3*\z,2.5) rectangle (3+3*\z+\x,2.5+\y) 
node[pos=.5] {$q,\neg p$ \\ $A_i = \{p\}$};
\node[scale=2, color=red] at (3+\x-3*\e+3*\z,2.5+\y-3*\e) {$\circlearrowleft$};
\node at (3+\x*.5+3*\z,2.5-2*\e) {$t^\dag$};

\draw[rounded corners, thick,]  (1.5+3*\z,0) rectangle (1.5+\x+3*\z,0+\y)
node[pos=.5] {$p$ \\ $A_i = \{p\}$};
\node at (1.5+\x*.5+3*\z,0-2*\e) {$\w^*_1$};

\draw[thick,->,draw=red] (1.5+\x*.5,0+\y+\e) -- (1.5+\x*.5,2.15);

\draw[thick,->,draw=red] (1.5+\x*.5+3*\z,0+\y+\e) -- (0+\x*.66+3*\z,2.5-\e);
\draw[thick,->,draw=red] (1.5+\x*.5+3*\z,0+\y+\e) -- (3+\x*.33+3*\z,2.5-\e);
\draw[thick,<->,draw=red] (3*\z+\x+\e,2.5+\y*.5) -- (3+3*\z - \e,2.5+\y*.5);

\draw[thick,->, gray] (0+3*\z+\x*.5,2.5+\y+\e) to[out=160,in=20] node [pos=.5, below] {$f$} (1.5+\x*.5,2.5+\y+\e);
\draw[thick,->, gray] (3+3*\z+\x*.5,2.5+\y+\e) to[out=160,in=20] node [pos=.3, above] {$f'$} (1.5+\x*.5,2.5+\y+\e);
\draw[thick,->, gray] (1.5+3*\z-\e,0+.5*\y) -- node [pos=.5, below] {Id} (1.5+\x+\e,0+.5*\y);

\end{tikzpicture}
\caption{A visual representation of Example~\ref{ex:oneagent}. The states on the left are the initial state space, $M$, and those on the right are $M^*$. The red, unlabeled arrows indicate $i$'s accessibility
  relation. The gray arrows indicate the 
  relation $T$, labeled according to the replacement scheme.} 
\label{fig: ex2}
\end{figure*}
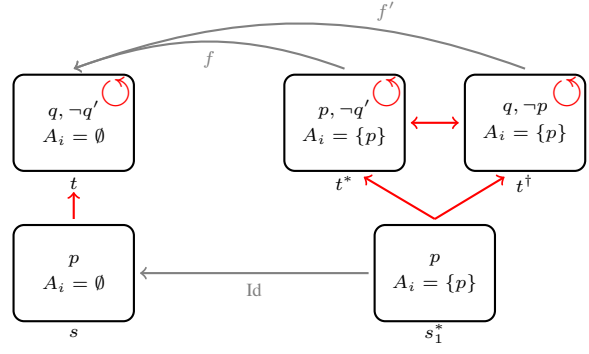

The following example explores the mechanics of the single agent transition rule and is presented diagrammatically in Figure \ref{fig: ex2}.

\begin{example}
\label{ex:oneagent}
Consider  an agent $i$ who is
uncertain how to interpret a novel proposition. Let $M$ be the initial
model with a state space $S = \{s,t\}$, where
$\L(s) = \{p\}$ and $\L(t) = \{q,q'\}$.  Think of $p$ as a real
proposition and $q$ and $q'$ as shadow propositions. In $s$, $p$ is
true, and in $t$, $q$ is true and $q'$ false. The agent is unaware of
$p$ and considers only $t$ possible: $\A_i(s) = \A_i(t) =
\emptyset$ and $\K_i(s) = \K_i(t) = \{t\}$. Obviously, the agent
places probability 1 on $t$. 

For some $\alpha \in [0,1]$, consider the transition rule
$\tau^\alpha$ such that $\tau^\alpha((M,s),p,i) = (M^*,s^*)$.   $M^*$ has the
state space $S^* = \{s^*,t^*, t^\dag\}$. Let $f$ be the $p$-replacement
scheme given by $f: p \mapsto q$ and let $f'$ be the $p$-replacement scheme
given by $f': p \mapsto q'$. Then $T = \{(s,s^*,id),
(t,t^*,f), (t,t^{\dag},f')\}$. In all states, $i$'s awareness is exactly
$p$, and in all states, she  considers both $t^*$ $t^\dag$ possible,
that is, 
$\K_i(s^*) = \K_i(t^*) = \K_i(t^\dag) = \{t^*,t^\dag\}$. 
%
Finally, $i$ puts probability 
$\alpha$ on state $t^*$ and $(1-\alpha)$ on state $t^\dag$. When $\alpha
= 1$, $i$ is sure that the novel proposition $p$ is what she was
representing by the shadow proposition $q$, and when $\alpha = 0$ she
is sure it was represented by $q'$. In between, the agent is uncertain
about the interpretation of the novel proposition. Notice this
uncertainty is not captured in $M$, as the agent is
unaware of $p$ in $M$, hence can not directly reason about its likelihood of
being represented by $q$ or $q'$. 
\qed
\end{example}

\subsection{Dynamics: The Multiagent Case}

In this section, we present a transition rule for multiagent
models. We assume that other agents do not realize that $i$ has
become aware, so their beliefs remain invariant under the model
transition. Although this is conceptually simple, it complicates the
definition of an acceptable rule because
the updated model must contain additional
states to handle other agents' beliefs. 
%
For example, if there are two agents, and $j$ initially (correctly) believes $i$ is unaware of $p$, then after $i$ becomes aware of $p$, $j$ must consider a state where $i$ is unaware of $p$, even though he no longer is. Furthermore, since $j$ still believes $i$ is unaware of $p$, the worlds that $i$ considers from the worlds that $j$ considers (i.e., $\K_i(\K_j(s))$) cannot contain $p$ in language, requiring yet more states to be added.

The following definition generalizes Definition~\ref{def:single} to models with multiple agents.

\begin{definition}
\label{def:multi}
$\tau: \S(\I,\Phi) \times \LQXAn(\Phi) \times \I \to  
\S(\I,\Phi)$ is an \emph{acceptable transition rule} if for all
$(M,s) \in \S(\I,\Phi)$,
$\phi \in \LQXAn(\Phi)$, and $i \in \I$, if
$\prop(\phi)\in \L(s)$,
$M = (\W, \val, \K_1,
  \ldots, \K_n, \A_1, \dots, \A_n, \PR_1, \ldots, \PR_n,\L)$, 
  $\tau((M,\w),\phi, i) = (M^*,\w^*)$, and
  $M^* = (\W^*, \val^*, \K_1^*,
  \ldots, \K_n^*, \A_1^*, \dots, \A_n^*, \PR_1^*, \ldots,
  \PR_n^*,\L^*)$,
  then either $\prop(\phi) \subseteq \A_i(s)$ and $(M,s)=(M^*,\w^*)$, or
  $\prop(\phi) \not\subseteq \A_i(s)$ and 
  there exists a relation $T \subseteq \W \times \W^* \times \RS(\phi)$
    such that the following hold:

    \begin{description}
\item[{\rm T1.}]
  For all $t^* \in \W^*$, there exists a unique $t \in \W$ and $f$ such
that $(t,t^*, f) \in T$; moreover,
 $t^*$ is an $f$-replacement of $t$. 
\item[{\rm T2.}] $(s,s^*,id) \in T$, $\A^*_i(\w^*) = \A_i(\w) \cup
\prop(\phi)$, and $\A^*_j(\w^*) = \A_j(\w)$ for $j \neq i$.
\item[{\rm T3.}]  
  If $(t,t^*,f) \in T$, $t \in \K_i(\w) \cup \{\w\}$,
  either $f \ne id$ or $(t^*,f) = (\w^*,id)$, and 
  $\PR_i(\w)(\cons(M,\phi,i)) \ne 0$, then the following
  conditions hold:
  \begin{description}
  \item[{\rm (a)}] $\K_i^*(t^*) = \{t^{\dag}: (t',t^{\dag},f') \in
T, \mbox{ for some } t' \in \K_i(\w) \cap \cons(M,\phi,i), f' \in \RS(\phi,i,\w)\}$. 
      \item[{\rm (b)}] If $t' \in \K_i(\w) \cap \cons(M,\phi,i)$, let
                $\rep(t') = \{t^{\dag} \in \K_i^*(t'): \exists f' 
                ((t',t^{\dag},f') \in T)\}$.  Then  
        $\PR_i^*(t^*)(\rep(t')) =
        \PR_i(\w)(t')/\PR_i(\w)(\cons(M,\phi,i))$. 
 \item[{\rm (c)}] $f(\A_{j}(t^*)) = \A_{j}(t)$ for all $j \neq i$.
  \end{description}
\item[{\rm T4.}] If $(t,t^*,f) \in T$,  then the following hold for all agents $j \ne i$, 
    and for $j=i$ if $t \notin \K_i(\w) \cup \{\w\}$ or if $f = id$
    and $t^*\ne \w^*$:
   \begin{description}
        \item[{\rm (a)}] $f(\A_j(t^*)) = \A_j(t)$.
        \item[{\rm (b)}]
          For all $t' \in\K_j(t)$, there exists $t^\dag \ne \w^*$ such that
          $(t',t^\dag,f) \in T$, and 
          $\K_j^*(t^*) = \{t^{\dag}: (t',t^{\dag},f) \in
        T \mbox{ for some } t' \in \K_j(t), \ t^\dag \ne \w^* \}$. 
   \item[{\rm (c)}] $\PR_j^*(t^*)(\rep(t')\cap \K_i(t^*)) = \PR_j(\w)(t')$. 
\end{description}  
\end{description}

\end{definition}

T1 and T2 are the same as in Definition~\ref{def:single} with the caveat that 
the awareness of agents other than $i$ does not change. 
%
T3 is largely the same as before, except that it adds a
requirement handling the awareness of other agents and it does
not apply to all states. 
%
For a state $t^\dag$ that corresponds to $t'$ via $f$,
the awareness 
of each agent $j \ne i$ changes according to $f$;
if $i$ becomes aware of $p$, then $i$ replaces $f(p)$ with $p$, so if $j$
was aware of $f(p)$ 
in $t'$, then $j$ must be aware of $p$ in $t^\dag$---$p$ replaces $f(p)$ in
$\A_j(t^*)$. 

The reason T3 no longer applies to all states is we need to allow other agents to maintain their beliefs.
Note that states $t^\dag$ for which $(t',t^\dag,id) \in T$ are
\emph{not} in $\K_i(s^*)$, since $id \notin \RS(\phi,i,\w)$ if there
propositions that $i$ is not aware of in $\phi$ and thus are not under the scope of T3.
This is because
$\prop(\phi) \subseteq \Phi$, which is disjoint from $\Phi'$, so we
cannot have $id(\prop(\phi)-\A_i(s)) \subseteq \Phi'$.   We use
 states $t^\dag$ where $(t',t^\dag,id) \in T$ and $t' \in \K_i(\w)$ to 
capture other agents' beliefs about $i$.

Finally, T4 describes agents' beliefs and awareness except for $i$'s
belief and awareness in states in $\K_i(\w^*)$.  Because other agents
do not know about $i$'s increased
awareness, T4 
also handles the construction of $i$'s beliefs and awareness in states
that are considered possible by other agents.
 Roughly speaking, T4 says that if $(t,t^*,f)\in T$, then each agent's
awareness changes according to $f$.
The states that $j$ considers possible at a state
that is an $f$-replacement of $t$ are exactly the
$f$-replacements of $\K_j(t)$. So knowledge is unperturbed except as
required by replacing shadow propositions where needed. The final
condition of T4 mirrors this but for probabilistic assessments.

\begin{example}
\label{ex:knowledgechange}
Consider what happens when agent $i$ in the model $M$ from Example
\ref{ex:jaware} becomes aware of $p'$ in state $\w_1$. Let $\tau$ be
a transition rule 
that maps $(M,\w_1)$ to $(M^*, \w_1^*)$ with states $\W^* = \{\w^*_{1}, \w^*_{2i},
\w^*_{1j},\w^*_{2j},\w^*_{3j},\w^*_{2ij},\w^*_{3ij},\w^*_{1ij}\}$. Letting $f$ denote the $p'$-replacement that
%
maps $p'$ to $q$, $T$ consists of $(\w_1, \w^*_{1}, id)$,
$(\w_2, \w^*_{2i}, f)$, and $(\w_n, \w^*_{nj}, id)$ and $(\w_n,
\w^*_{nij}, f)\}$ for $n \in \{1,2,3\}$. 

The awareness functions in $M^*$ are defined as follows:
$\A_i^*(\w^*_1) = \A_j^*(\w^*_1) = \A_i^*(\w^*_{2i}) = 
\A_j^*(\w^*_{2i}) = \A_j^*(\w^*_{2j}) = \{p,p'\}$ and $\A_i^*(\w^*_{2j}) =
\A_i^*(\w^*_3) = \A_j^*(\w^*_3) = \{p\}$. The reachability functions
in $M^*$ are defined as follows: $\K_i^*(\w^*_1) =
\K_i^*(\w^*_{2i}) = \{\w^*_{2i}\}$; $\K_j^*(\w^*_1) = \K_j^*(\w^*_{2i}) =
\K_j^*(\w^*_{2j}) = \{\w^*_{2j}\}$; $\K_i^*(\w^*_{2j}) = \K_i^*(\w^*_{3}) =
\{\w^*_{2j}, \w^*_3\}$; and  $\K_j^*(\w^*_3) = \{\w^*_3\}$.

At $\w^*_1$, $i$ believes that he is fully aware and
also knows that $p$ is true. This is because he has disregarded the
possibility of $\w_3$, since $\w_3 \notin \cons(p',i)$. Thus, despite
fact that $i$ only became aware of $p'$, and did not directly learn the
truth of any proposition, he has implicitly learned that $p$
holds. 
Moreover, notice that $j$ believes $i$ is unaware of $p'$.
\qed
\end{example}


The transition rule in Example~\ref{ex:knowledgechange}, as opposed to
that in Example~\ref{ex:oneagent}, adds many more states. This is
because when agent $i$ becomes aware of $p'$, agent $j$ does not know
this happened. Thus, we need a set of states that represent $j$'s (now
incorrect) beliefs that $i$ is unaware of $p'$: these are the states
$\w^*_{1j}$, $\w^*_{2j}$, and $\w^*_{3j}$. In addition, $i$ knows that
$j$ does not know she became aware of $p'$. So we also have to add more
states to capture $i$'s correct beliefs about $j$'s incorrect beliefs
about $i$'s awareness; these are the states $\w^*_{1ij}$,
$\w^*_{2ij}$, and $\w^*_{3ij}$.

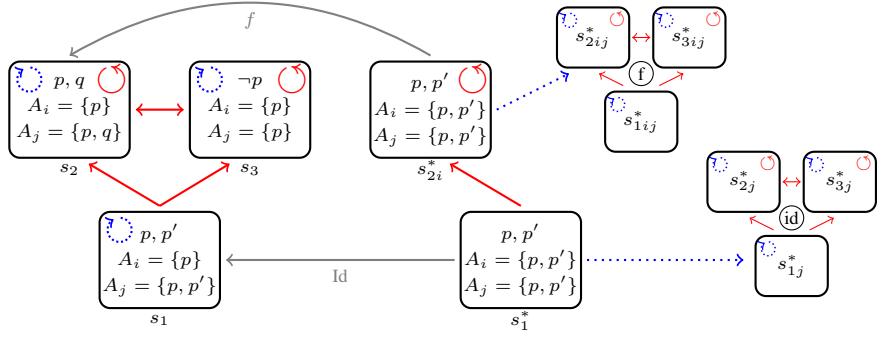
\begin{figure*}[]
\centering
\begin{tikzpicture}[scale=.8, font = \scriptsize,every text node part/.style={align=center}]
\def\x{2};
\def\y{1.6};
\def\e{.1};
\def\z{2};

\draw[rounded corners, thick,]  (0,2.5) rectangle (0+\x,2.5+\y) 
node[pos=.5] {$p,q$ \\ $A_i = \{p\}$ \\ $A_j = \{p,q\}$};
\node[scale=2, color=red] at (0+\x-3*\e,2.5+\y-3*\e) {$\circlearrowleft$};
 \draw [->,anchor=center, thick, densely dotted, blue]%
            (0+3.5*\e,2.5+\y-1*\e) arc (85:-240:.2);
\node at (0+\x*.5,2.5-2*\e) {$\w_2$};

\draw[rounded corners, thick,]  (0+3*\z,2.5) rectangle (0+3*\z+\x,2.5+\y) 
node[pos=.5] {$p,p'$ \\ $A_i = \{p,p'\}$ \\ $A_j = \{p,p'\}$};
\node[scale=2, color=red] at (0+\x-3*\e+3*\z,2.5+\y-3*\e) {$\circlearrowleft$};
\node at (0+\x*.5+3*\z,2.5-2*\e) {$\w^*_{2i}$};

\draw[rounded corners, thick,]  (3,2.5) rectangle (3+\x,2.5+\y)
node[pos=.5] {$\neg p$ \\ $A_i = \{p\}$ \\ $A_j = \{p\}$};
\node[scale=2, color=red] at (3+\x-3*\e,2.5+\y-3*\e) {$\circlearrowleft$};
 \draw [->,anchor=center, thick, densely dotted, blue]%
           (3+3.5*\e,2.5+\y-1*\e) arc (85:-240:.2);
\node at (3+\x*.5,2.5-2*\e) {$\w_3$};

\draw[rounded corners, thick,]  (1.5,0) rectangle (1.5+\x,0+\y)
node[pos=.5] {$p,p'$ \\ $A_i = \{p\}$ \\ $A_j = \{p,p'\}$};
 \draw [->,anchor=center, thick, densely dotted, blue]%
            (1.5+3.5*\e,0+\y-1*\e) arc (85:-240:.2);
\node at (1.5+\x*.5,0-2*\e) {$\w_1$};

\draw[rounded corners, thick,]  (1.5+3*\z,0) rectangle (1.5+\x+3*\z,0+\y)
node[pos=.5] {$p,p'$ \\ $A_i = \{p,p'\}$ \\ $A_j = \{p,p'\}$};
\node at (1.5+\x*.5+3*\z,0-2*\e) {$\w^*_1$};

\draw[thick,->,draw=red] (1.5+\x*.5,0+\y+\e) -- (3+\x*.33,2.5-\e);
\draw[thick,->,draw=red] (1.5+\x*.5,0+\y+\e) -- (0+\x*.66,2.5-\e);
\draw[thick,<->,draw=red] (0+\x+\e,2.5+\y*.5) -- (3-\e,2.5+\y*.5);

\draw[thick,->, dotted, draw=blue] (1.5+\x+3*\z+\e,0+.5*\y) -- (6.3+3*\z-\e,0+.5*\y);
\draw[thick,->,draw=red] (1.5+\x*.5+3*\z,0+\y+\e) -- (0+\x*.66+3*\z,2.5-\e);
\draw[thick,->, dotted, draw=blue] (0+\x+\e+3*\z,2.5+\y*.5) -- (3.6+3*\z-3*\e,3.1+\y*.5);

\def\sx{1.4};
\def\sy{1.2};
\coordinate (id) at (7+3*\z,.6+\y*.5);
\draw[] ($(id)+ (0,\e)$) circle (.2 cm) node {id};
\draw[rounded corners, thick,]  ($(id) + (2*\e,2*\e)$) rectangle ($(id) + (\sx,\sy)$)
node[pos=.5] {$\w^*_{3j}$};
 \draw [->,anchor=center, line width=0.2mm, densely dotted, blue]%
            ($(id) + (2*\e,\sy) + (2.2*\e,-1*\e)$) arc (85:-240:.115);

\node[scale=1, color=red] at ($(id) + (2*\e,\sy) + (2*\e,-2*\e) + (\sx,0) - (6*\e,0)$) {$\circlearrowleft$};
\draw[rounded corners, thick,]  ($(id) - (2*\e,-2*\e)$) rectangle ($(id) - (\sx, -\sy)$)
node[pos=.5] {$\w^*_{2j}$};
 \draw [->,anchor=center, line width=0.2mm, densely dotted, blue]%
            ($(id) + (- \sx,\sy) + (2.2*\e,-1*\e)$) arc (85:-240:.115);
\node[scale=1, color=red] at ($(id) + (- \sx,\sy) + (2*\e,-2*\e)+ (\sx,0) - (6*\e,0)$) {$\circlearrowleft$};
\draw[rounded corners, thick,]  ($(id) - (-\e,2*\e) - (.5*\sx,0)$) rectangle ($(id) - (-.5*\sx +\e, \sy)$)
node[pos=.5] {$\w^*_{1j}$};
 \draw [->,anchor=center, line width=0.2mm, densely dotted, blue]%
            ($(id) + (- \e ,0) - (.5*\sx,0) + (4.2*\e,-3*\e)$) arc (85:-240:.115);
            
\draw[->,draw=red] ($(id) + (3*\e,-\e)$) -- ($(id) + (.5*\sx,\e)$);
\draw[->,draw=red] ($(id) + (-3*\e,-\e)$) -- ($(id) + (-.5*\sx,\e)$);
\draw[<->,draw=red] ($(id) + (-1.5*\e,.5*\sx)$) -- ($(id) + (1.5*\e,.5*\sx)$);

\def\sx{1.4};
\def\sy{1.2};
\coordinate (f) at (4.5+3*\z,3+\y*.5);
\draw[] ($(f)+ (0,\e)$) circle (.2 cm) node {f};
\draw[rounded corners, thick,]  ($(f) + (2*\e,2*\e)$) rectangle ($(f) + (\sx,\sy)$)
node[pos=.5] {$\w^*_{3ij}$};
 \draw [->,anchor=center, line width=0.2mm, densely dotted, blue]%
            ($(f) + (2*\e,\sy) + (2.2*\e,-1*\e)$) arc (85:-240:.115);
\node[scale=1, color=red] at ($(f) + (2*\e,\sy) + (2*\e,-2*\e) + (\sx,0) - (6*\e,0)$) {$\circlearrowleft$};
\draw[rounded corners, thick,]  ($(f) - (2*\e,-2*\e)$) rectangle ($(f) - (\sx, -\sy)$)
node[pos=.5] {$\w^*_{2ij}$};
 \draw [->,anchor=center, line width=0.2mm, densely dotted, blue]%
            ($(f) + (- \sx,\sy) + (2.2*\e,-1*\e)$) arc (85:-240:.115);
\node[scale=1, color=red] at ($(f) + (- \sx,\sy) + (2*\e,-2*\e)+ (\sx,0) - (6*\e,0)$) {$\circlearrowleft$};

\draw[rounded corners, thick,]  ($(f) - (-\e,2*\e) - (.5*\sx,0)$) rectangle ($(f) - (-.5*\sx +\e, \sy)$)
node[pos=.5] {$\w^*_{1ij}$};
 \draw [->,anchor=center, line width=0.2mm, densely dotted, blue]%
            ($(f) + (- \e ,0) - (.5*\sx,0) + (4.2*\e,-3*\e)$) arc (85:-240:.115);
            
\draw[->,draw=red] ($(f) + (3*\e,-\e)$) -- ($(f) + (.5*\sx,\e)$);
\draw[->,draw=red] ($(f) + (-3*\e,-\e)$) -- ($(f) + (-.5*\sx,\e)$);
\draw[<->,draw=red] ($(f) + (-1.5*\e,.5*\sx)$) -- ($(f) + (1.5*\e,.5*\sx)$);

\draw[thick,->, gray] (0+3*\z+\x*.5,2.5+\y+\e) to[out=150,in=30] node [pos=.5, below] {$f$} (0+\x*.5,2.5+\y+\e);
\draw[thick,->, gray] (1.5+3*\z-\e,0+.5*\y) -- node [pos=.5, below] {Id} (1.5+\x+\e,0+.5*\y);

\end{tikzpicture}
\caption{A visual representation of Example
  \ref{ex:knowledgechange}. The red arrows indicate $i$'s accessibility
  relation; the blue (dotted) arrows indicate $j$'s accessibility relation. The
  gray arrows indicate the 
  relation $T$, labeled according to the replacement scheme.
The two sets of three states on the right side of the figure are
  both copies of the original three states, under different
    replacement schemes; the upper three states are $f$-replacements and
  the lower three are $id$-replacements.}  
\label{fig: ex3}
\end{figure*}

\subsection{Dynamics: Syntax and Axioms}\label{sec:axioms}

In this section we briefly explore an extension of the language
presented in Section~\ref{sec:lang} that allows us to capture the
dynamic aspects of our transition rule. We then present some axioms
which we hope will help convey the requirements of a valid transition
rule more directly. While these axioms are all sound, getting a
complete axiomatization 
seems to require more effort.  We are currently exploring this.

Given that we are interested in what happens after an agent becomes of a
formula, it seems reasonable to consider extending the language by
adding formulas of the form $[\phi,i]\psi$ to the underlying language
$\LQXAn$ considered in this paper, where $[\phi,i]\psi$ is interpreted
as ``after $i$ becomes aware of $\phi$, $\psi$ is true.''
Such a formula is interpreted relative to a probabiistic awarness
structure $M$ and a transition rule $\tau$ in the obvious way:
$$\mbox{$(M,\w, \tau)\sat [\phi,i] \psi$ if $((\tau((M,\w),
\phi,i),\tau) \sat \psi$.}$$ 
Note that this definition allows for occurrences of formulas of
the form $[\phi',j]\psi'$ in $\psi$ (although not in $\phi$).

In this language, we can capture many of the properties of the
transition function $\tau$.  For example, the following two axioms are sound:
\begin{description}
\item[{\rm A$^{\!*}$.}] $A_i \psi \iff [\phi,i] A_i(\phi \land
  \psi)$ if $\prop(\psi) \cap \prop(\phi) = \emptyset$.
\item[{\rm AK.}]
 $K_j\psi \land A_j\psi \iff [\phi,i] (K_j\psi \land  [\phi,i] A_j\psi$). 
\end{description}
A$^{\!*}$ states that after becoming aware of $\phi$ agent $i$ is aware of
$\phi$ and everything that he was initially aware of.
AKo states that the knowledge and awareness of other agents
remains invariant.

To  capture how an agent's beliefs changes as a result of an update,
we need to be able to talk about consistency.
Let $\Consis(\phi,i)$ be a new formula that is true at a
state $t$ if $t$ is consistent with $i$ becoming aware of $\phi$.  
If $\prop(\phi)$ has no primitive propositions, then $\Consis(\phi,i)$
is equivalent to ${\mathit true}$, and if 
$|\prop(\phi)| = 1$, then $\Consis(\phi,i)$ is equivalent
to $\exists x \neg A_i(x)$.  But if $|\prop(\phi)| > 1$, then it can
be shown that $\Consis(\phi,i)$ is not equivalent to a formula in
$\LQXAn$.   Using $\Consis(\phi,i)$, we can provide some
sound axioms that capture how an agent's beliefs change:
\begin{description}
\item[{\rm Ka.}] $(\ell_i(\Consis(i,\phi) > 0) \land
  K_i(\Consis(i,\phi) \rimp \psi)) \rimp [\phi,i] K_i\psi$. 
\item[{\rm Kb.}] $(\ell_i(\Consis(i,\phi) > 0 \land [\phi,i] K_i\psi[x/\psi']) \rimp K_i (\Consis(\phi,i)
  \rimp \exists x \psi)$.
\item[{\rm Pra.}] $\ell_i(\psi \land \Consis(\phi,i))  > \alpha
  \ell_i(\Consis(\phi,i)) \rimp   [\phi,i](\ell_i(\psi) > \alpha)$. 
\item[{\rm Prb.}] $(\ell_i(\Consis(i,\phi) > 0 \land [\phi,i]
  (\ell_i\psi[x/\phi]) > \alpha) \rimp 
 \ell_i( \exists x \psi \land \Consis(\phi,i)) > \alpha
  \ell_i(\Consis(\phi,i))$.
\end{description}
Ka states that
after becoming aware of $\phi$, $i$ knows everything he knew before he
was aware of $\phi$ that was consistent with becoming aware of $\phi$,
provided that 
becoming aware of $\phi$ has positive probability.  Kb is almost
a converse.  It says that if after becoming aware of $\phi$ $i$ knows
$\psi$, then before becoming aware,  $i$ knew that $\psi$ (with
occurrences of $\phi$ replaced by an existential) held in all states
where it was consistent for $i$ to become aware of $\phi$.  
Pra and Prb are analogues of Ka and Kb for probability formulas.  Note
that $\ell_i(\psi \land \Consis(\phi,i))  > \alpha
  \ell_i(\Consis(\phi,i))$ essentially says that the probaiblity of
  $\psi$ conditional on $\Consis(\phi,i)$ is greater than $\alpha$.
Since we do not have conditional probability formulas in the language,
we cannot say this directly.

\section{Information Disclosure}
\label{sec:info}
\def\signal{rating}

The following section is a simple application of our model to
the problem of information disclosure. In the interest of space and
simplicity, we consider a highly stylized setting:
There are two agents, a buyer and a seller (we can think of the seller
as a restaurateur and the buyer as a potential patron). The agents'
belief and awareness will be modeled via a probabilistic awareness
structure for 2 agents along with a transition rule. We can think of
the agents as playing 
a game (i.e., they have actions and payoffs associated with these
actions), but the game is not described explicitly in the epistemic model.

The initial state space (i.e., before the buyer become aware of a
formula of which he was previously unaware) is $\{s_n^m,t_n^m\}_{(n,m)
  \in \{1,\dots, N\} \times \{1, \ldots,   M\}}$. The lower index ($n$)
represents the \emph{type} or quality 
of the seller. So in state $s_n^m$ or $t_n^m$, the seller has
quality $n$; we assume that the buyer's value for a purchase is
increasing in $n$ (i.e., the buyer gets better-quality meals, which he
values more highly, at a
better-quality restaurant). The upper index ($m$) represents the
buyer's type: how the buyer
reacts to becoming aware of new features (explained in detail shortly). 

The $s$ states (i.e., states of the form $s_n^m$ for some $n$ and $m$)
are ``real'' in the sense that they contain no shadow
propositions. The seller, who we assume is fully aware, considers only
$s$ states possible. The $t$ states represent the buyer's
understanding of the world, containing shadow propositions in place of
the real propositions that he is unaware of but exist in the $s$
states.  We assume that (it is commonly
known that) the seller knows
his type $n$ but not how the buyer will react to information, so
$\K_s(s_n^m) \subseteq \{s_n^{m'}: m' \in \{1, \ldots ,
M\}\}$. On the other hand, (it is commonly known that) the buyer knows
how he will react to information, but does not know the seller's type,
so
$\K_b(s_n^m) =  \K_b(t_n^m) = \{ t_{n'}^m :
n' \in \{1, \ldots, N\}\}$.

A list of propositions $\{p_1, \ldots, p_K\}$ is a \emph{\signal{} structure},
and the $p_k$'s are called \emph{\signal{}s}, if (1)
exactly one proposition $p_k$ is true at each state $s_n^m$ and (2)
if $p_k$ is true at $s_n^m$, $p_{k'}$ is true at $s_{n'}^{m'}$, and $k
> k'$, then $n > n'$.
Thus, higher \signal{}s indicate higher-quality sellers.
Note that this condition implies that the same \signal\ must
hold at $s_n^m$ and $s_n^{m'}$: the seller knows the \signal.
In general, a \signal{} may be true at more than one state; a
\signal{} $p$ is \emph{perfectly 
    informative}  if there is exactly one $n$ such that $p$ is
true of only states of the form $s_n^m$.
We assume that the seller is aware of the rating, but the buyer is
not.  Formally, we assume a particular \signal\ structure, given by
$p_1, \ldots, p_K$, such that the seller is aware of $p_1, \ldots, p_K$
at all states of the form $s_n^m$.  The buyer is not aware of these
formulas at any state.  In some states of the form $t_m^n$ there is no
rating structure in the language (these are the states where it does
not occur to the buyer that there is a rating structure).  In other
states there of the form $t_n^m$, there are one or more rating
structures, but these are represented by shadow propositions $x_1,
\ldots, x_{K'}$; the buyer is not aware of these shadow propositions,
but understands that the seller knows the actual rating structure
(so, in these states, formulas of the form $\neg A_b(x_j) \land A_s(x_j)
\land (K_s(x_j) \lor K_s(\neg x_j)$ hold).  
We assume that the superscript $m$ in $t^m_n$ indicates which rating
structures are in the language at $t^m_n$, so if 
$\{x_1, \ldots, x_K\}$ is a rating structure
and $\{x_1, \ldots, x_K\} \subseteq \L(t^m_n)$, then $\{x_1
\ldots x_K\} \subseteq \L(t^m_{n'})$ for all $n'$.

The seller sets a price and
possibly discloses his \signal{};
the buyer then decides whether to buy (i.e., whether to eat at the
restaurant).  Before deciding, the buyer updates his beliefs based
on whatever information the seller discloses and his beliefs about the
seller's.
If the buyer was unaware of $p_k$ and $p_k$ is
disclosed, then the buyer first updates his beliefs according to the
transition rule to become aware of $p_k$, then he conditions on the
fact that $p_k$ is true.
Among other things, the index $m$ captures how the buyer
will update when she becomes aware of a rating (i.e., it encodes
a replacement function).
An equilibrium of this game consists of (1) an action for each type of
seller, and (2) beliefs and an action for the buyer, for each action of the
seller, where (1) and (2) are such that both parties are maximizing
(the expected revenue and  the
expected value of the
purchase, respectively).  

\commentout{
First, we analyze the (standard) case where the buyer is aware of
$\{p_1, \ldots, p_K\}$ from the beginning. 
Standard arguments apply to indicate that all types of seller will
disclose their \signal{} (except possibly those types with the lowest \signal{}, $p_1$) and the seller will charge the
expected value of purchase, conditional on the \signal{}.
Sellers who have the
highest \signal{}, $p_K$,  
would certainty reveal. Thus, a buyer that does not see any \signal{} will
believe that the restaurant is of at most the second-highest \signal{}, $p_{K-1}$,
but possibly less than of second-highest.  To prevent the
buyer from thinking that his restaurant is of even lower quality, a
buyer with the second-highest \signal{} will reveal that too,
and so on.
}

As we observed in the introduction, if the buyer is aware of the
rating structure, then all sellers (except possibly those with the
lowest rating) will reveal their rating.
On the other hand, if the buyer is unaware of the \signal{} structure $\{p_1,
\ldots, p_K\}$, we must explicitly indicate how he would treat the lack
of revelation.  This is the role of the buyer's type.
We assume here that when no \signal{} is revealed,
the buyer's beliefs do not change at all: the buyer continues to be
unaware of the \signal{}.  But note that this is fundamentally
different from the buyer knowing 
that the \signal{} could have been disclosed but was not. 
Under this assumption, the unraveling argument given in the
introduction fails: any type 
with a \signal{} imparting lower-than-average quality would prefer the
buyer to retain his original belief; not disclosing does not have
any effect on these beliefs, since the buyer's unawareness prevents him from
reasoning about the actions of other types.

Given the immediate failure of the unraveling argument, the more
interesting aspect of this model is the dependency of the seller's
beliefs on the \emph{transition rule}, as encoded by the $m$ index.
The key point is that $s_n^m$ and $s_n^{m'}$ are distinguished by
 the buyer's reaction to becoming aware of a new \signal{}.
%
 We provide an example showing how the seller's beliefs about the
 transition rule
  determine whether the seller decides to disclose his rating.

Suppose that $N = 3$, so that there are three qualities of seller
(restaurant) and $M =
2$. Suppose that the buyer's utility for a purchase if the restaurant has
quality $n$ is $n$, and that at $s_n^m$, the buyer initially believes
that $t_1^m$, $t_2^m$, and $t_3^m$ have probability 
$\tfrac12$, $\tfrac14$, and $\tfrac14$, respectively.  
This indicates that in the absence of any
signal, when the buyer does not update his beliefs at all, his
expected utility is $1 \times \tfrac12 + 2 \times \tfrac14 + 3 \times
\tfrac14 = \tfrac74$.  

Suppose that the \signal{} $p_n$ is true at state $s_n^m$, for $n = 1, 2,
3$, so $p_n$ is perfectly informative; $\L(s_n^m) = \{p_1, p_2,
p_3\}$.  We take $\L(t_m^n) = \{x_1, x_2, x_3, y_1\}$, where
these are all shadow \signal{}s.  Think of $x_n$ as
the shadow signal corresponding to $p_n$, so that
$x_n$ is true only at the states $t_n^m$ for $n=1,2,3$.   The shadow
\signal{} $y_1$ is a coarsening of $x_1$ and $x_2$; it is true at both
$t_1^m$ and $t_2^m$. 
  
The transition rule is such that, if the buyer becomes aware of $p_n$, the
situation $(M, s^m_n)$ is mapped to the situation $(M^*_n,
s^{m^*}_n)$ with relation $T_n$  
containing $(t^1_{n'}, t^{1*}_{n'}, f^1_n)$ and $(t^2_{n'}, t^{2*}_{n'}, f^2_n)$ for $n' \in \{1,2,3\}$, where the replacements are given by
$f^1_n: p_n \mapsto x_n$ for $n \in \{1,2,3\}$, and  $f^2_1: p_1 \mapsto y_1$, 
$f^2_2: p_2 \mapsto y_1$, and $f^2_3: p_3  \mapsto x_3$.
That is, in states of the form $t^1_n$, the buyer interprets $p_k$ as arising
from the perfectly informative shadow \signal{} structure
$\{x_1,x_2,x_3\}$, and in states $t^2_n$, he interprets it as arising
from the coarser shadow \signal{} structure $\{y_1,x_3\}$, conflating
$p_1$ and $p_2$. 

As always, the highest type, a seller in state $s^m_3$, will disclose
regardless of his beliefs about the transition rule. 
Conversely, a seller in state $s^m_1$ never has a reason to
disclose. However, a seller in state $s^m_2$ will disclose only when
he is sufficiently confident the seller will correctly interpret
$p_2$ as being perfectly informative. In state $s^1_2$, disclosing imparts
beliefs 
(in the updated model) that places probability 1 on state $t^1_2$, and
a willingness to pay of $2 > \tfrac74$. Thus, if he knew that he was in
state $\w^1_2$, he would certainly want 
to disclose $p_2$.  If, however, he knew he was in state $s^2_2$,
disclosing imparts a beliefs (in the updated model) that
places probability $\tfrac23$ on state $t^1_1$ and $\tfrac13$ on state
$t^1_2$, and a willingness to pay of $\tfrac43 <
\tfrac74$.  

\section{Discussion}
\label{sec:conc}

This paper develops a  modal logic that allows agents to reason both
about probabilities and about their own and others' awareness. We
introduce the semantic notion of a model transition, which captures
how a model changes when an agent becomes aware of a new formula. 
%
In contrast to prior work on the subject, our transition rule allows
the relative likelihood of two formulas 
that an agent was aware of to change arbitrarily after she becomes
more aware. This is because the agent can, to some extent, reason
about her own unawareness. If, for example, after becoming aware of
$\phi$ she believes that having been unaware of $\phi$ was correlated
with the truth of some other formula $\psi$, then her probabilistic
assessment of $\psi$ can change. 

We then apply this model to a simple economic environment of
information disclosure, where one agent can strategically decide to
make another agent aware of a new formula. We show that the agent's
beliefs about how others will react to novel information (i.e., her
beliefs about the model transition rule) determine her decision to
disclose information and expand the awareness of other agents. 

In future work, we hope to explore further applications of dynamic
awareness.  We believe that thinking about how awareness changes will
prove critical in understanding economic puzzles and the general
problem of updating in the presence of unawareness.  
We would also hope to provide a sound and complete axiomatic
characterization of our transition rule, in the spirit of the
discussion in Section~\ref{sec:axioms}.

\section*{Acknowledgments} Halpern was supported in part by
NSF grants IIS-178108 and IIS-1703846, a grant from the Open
Philanthropy Foundation, ARO grant W911NF-17-1-0592, and MURI grant
W911NF-19-1-0217.

\bibliographystyle{kr.bst}
\bibliography{joe,PA,z}

\begin{thebibliography}{}

\bibitem[\protect\citeauthoryear{Board and Chung}{2009}]{BC09}
Board, O., and Chung, K.-S.
\newblock 2009.
\newblock Object-based unawareness: theory and applications.
\newblock Working paper 378, University of Pittsburgh.

\bibitem[\protect\citeauthoryear{Cox, Serv{\'a}tka, and
  Vadovi{\v{c}}}{2017}]{cox2017status}
Cox, J.~C.; Serv{\'a}tka, M.; and Vadovi{\v{c}}, R.
\newblock 2017.
\newblock Status quo effects in fairness games: reciprocal responses to acts of
  commission versus acts of omission.
\newblock {\em Experimental Economics} 20(1):1--18.

\bibitem[\protect\citeauthoryear{Esponda and
  Vespa}{2014}]{esponda2014hypothetical}
Esponda, I., and Vespa, E.
\newblock 2014.
\newblock Hypothetical thinking and information extraction in the laboratory.
\newblock {\em American Economic Journal: Microeconomics} 6(4):180--202.

\bibitem[\protect\citeauthoryear{Fagin and Halpern}{1988}]{FH}
Fagin, R., and Halpern, J.~Y.
\newblock 1988.
\newblock Belief, awareness, and limited reasoning.
\newblock {\em Artificial Intelligence} 34:39--76.

\bibitem[\protect\citeauthoryear{Fagin and Halpern}{1994}]{FH3}
Fagin, R., and Halpern, J.~Y.
\newblock 1994.
\newblock Reasoning about knowledge and probability.
\newblock {\em Journal of the ACM} 41(2):340--367.

\bibitem[\protect\citeauthoryear{Fagin \bgroup et al\mbox.\egroup
  }{1995}]{FHMV}
Fagin, R.; Halpern, J.~Y.; Moses, Y.; and Vardi, M.~Y.
\newblock 1995.
\newblock {\em Reasoning About Knowledge}.
\newblock Cambridge, MA: MIT Press.
\newblock A slightly revised paperback version was published in 2003.

\bibitem[\protect\citeauthoryear{Halpern and R\^ego}{2013}]{HR09}
Halpern, J.~Y., and R\^ego, L.~C.
\newblock 2013.
\newblock Reasoning about knowledge of unawareness revisited.
\newblock {\em Mathematical Social Sciences} 66(2):73--84.

\bibitem[\protect\citeauthoryear{Heifetz, Meier, and Schipper}{2006}]{HMS03}
Heifetz, A.; Meier, M.; and Schipper, B.
\newblock 2006.
\newblock Interactive unawareness.
\newblock {\em Journal of Economic Theory} 130:78--94.

\bibitem[\protect\citeauthoryear{Jin, Luca, and Martin}{2018}]{jin2018no}
Jin, G.~Z.; Luca, M.; and Martin, D.
\newblock 2018.
\newblock Is no news (perceived as) bad news? {A}n experimental investigation
  of information disclosure.
\newblock Technical report, National Bureau of Economic Research.

\bibitem[\protect\citeauthoryear{John, Barasz, and
  Norton}{2016}]{john2016hiding}
John, L.~K.; Barasz, K.; and Norton, M.~I.
\newblock 2016.
\newblock Hiding personal information reveals the worst.
\newblock {\em Proceedings of the National Academy of Sciences}
  113(4):954--959.

\bibitem[\protect\citeauthoryear{Karni and Vier{\o}}{2013}]{karni2013reverse}
Karni, E., and Vier{\o}, M.-L.
\newblock 2013.
\newblock ``{R}everse {B}ayesianism'': A choice-based theory of growing
  awareness.
\newblock {\em American Economic Review} 103(7):2790--2810.

\bibitem[\protect\citeauthoryear{Modica and Rustichini}{1994}]{MR94}
Modica, S., and Rustichini, A.
\newblock 1994.
\newblock Awareness and partitional information structures.
\newblock {\em Theory and Decision} 37:107--124.

\bibitem[\protect\citeauthoryear{Modica and Rustichini}{1999}]{MR99}
Modica, S., and Rustichini, A.
\newblock 1999.
\newblock Unawareness and partitional information structures.
\newblock {\em Games and Economic Behavior} 27(2):265--298.

\bibitem[\protect\citeauthoryear{Piermont}{2019}]{piermont2019unforeseen}
Piermont, E.
\newblock 2019.
\newblock Unforeseen evidence.
\newblock {\em arXiv preprint arXiv:1907.07019}.

\bibitem[\protect\citeauthoryear{Sillari}{2008}]{Sil08}
Sillari, G.
\newblock 2008.
\newblock Quantified logic of awareness and impossible possible worlds.
\newblock {\em Review of Symbolic Logic} 1(4):514--529.

\end{thebibliography}

\end{document}